\pdfoutput=1
\documentclass[11pt]{article}
\usepackage{tabularx}
\usepackage{hyperref}
\usepackage{emnlp2021}
\usepackage{graphicx}
\usepackage{CJKutf8}
\usepackage{amssymb}
\usepackage{amsmath}
\usepackage{times}
\usepackage{latexsym}
\usepackage{bbding}
\usepackage[T1]{fontenc}
\usepackage[utf8]{inputenc}

\usepackage{microtype}

\title{A Chinese Multi-type Complex Questions Answering Dataset over Wikidata}
 \author{Jianyun Zou$^{\dag}$ \and Min Yang$^{\ddag}$ \and Lichao Zhang$^{\dag}$ \and Yechen Xu$^{\dag}$ \\
 {\bf Qifan Pan}$^{\dag}$ \and  {\bf Fengqing Jiang}$^{\dag}$ \and {\bf Ran Qin}$^{\dag}$ \and {\bf Shushu Wang}$^{\dag}$ \\
 {\bf Yifan He}$^{\#}$ \and {\bf Songfang Huang}$^{\#}$ \and {\bf Zhou Zhao}$^{\dag}$ \\
         Zhejiang University$^{\dag}$\\
         Shenzhen Institute of Advanced Technology, Chinese Academy of Sciences$^{\ddag}$ \\ Alibaba Group$^{\#}$}
\begin{document}
\begin{CJK*}{UTF8}{gbsn}
\maketitle
\begin{abstract}

Complex Knowledge Base Question Answering is a popular area of research in the past decade.
Recent public datasets have led to encouraging results in this field, but are mostly limited to English and only involve a small number of question types and relations, hindering research in more realistic settings and in languages other than English. In addition, few state-of-the-art KBQA models are trained on
Wikidata, one of the most popular real-world knowledge bases.

We propose CLC-QuAD, the first large scale complex Chinese semantic parsing dataset over Wikidata to address these challenges. Together with the dataset, we present a text-to-SPARQL baseline model, which can effectively answer multi-type complex questions, such as factual questions, dual intent questions, boolean questions, and counting questions, with Wikidata as the background knowledge. We finally analyze the
performance of SOTA KBQA models on this dataset and identify the challenges facing Chinese KBQA.

% Considering Wikidata as the background KG, there are still limited question answer systems based it and it accepts a wide variety of corresponding SPARQL query for multi-types complex questions.
% In this paper, we propose a text-to-SPARQL model for answering multi-types complex questions directly like fact question, dual intent question, boolean question and counting question.
% In addition, to encourage Chinese complex KBQA development we propose a large scale complex KBQA dataset, CLC-QuAD.
% We conduct experiments on LC-QuAD 2.0 and CLC-QuAD dataset both.
% Our experiments demonstrate that the proposed method achieve state-of-the-art compared with methods transferring from other dataset. And we reveal several corresponding SPARQL components accuracy between different types of questions.
% Furthermore, the performance in CLC-QuAD is inferior than LC-QuAD 2.0, showing that Chinese complex KBQA is a challenging task and the new dataset is released together with the model to promote more research in this direction.
\end{abstract}

\section{Introduction}
Knowledge Base Question Answering (KBQA) aims at answering factoid questions based on a knowledge base (KB) and
has attracted considerable attention due to its popular downstream applications ~\cite{FREE917,WebQuestions,WebQuestionsSP,LC-QUAD}. 
Previous KBQA methods have achieved remarkable progress when dealing with simple question types and relations. For example, current state-of-the-art methods \cite{STAGG,AMPCNN,improve_relation_detect} on the SimpleQuestion dataset \cite{Simple_Question}, which consists of simple questions and corresponding facts from Freebase, have achieved 90\% accuracy.
However, real-world KBQA applications often involve multi-type complex questions (e.g., fact questions, dual intent questions, boolean questions and counting questions), which are underexplored by previous KBQA methods. 

%The previous methods are mostly limited in the number of question types and the scale of knowledge graph
%Because of its popular downstream applications, it has attracted many researchers’ attentions ~\cite{FREE917,WebQuestions,WebQuestionsSP,LC-QUAD}. When question types and relations are simple, popular algorithms~\cite{STAGG,AMPCNN,improve_relation_detect} can obtain satisfactory results: on SimpleQuestion \cite{Simple_Question}, a dataset consisting of questions and corresponding facts from Freebase, aforementioned algorithms achieve over 90\% execution accuracy. Therefore, recently works have gradually shifted focus to complex KBQA. 

Recently, several approaches have been proposed for answering complex questions \cite{ACL_CATT_KBQA,QALD-9,emnlp_state-transition,AQG_RANKING_MODEL}. Despite the effectiveness of previous studies, there are two major limitations for answering multi-type complex questions in practice. First, 
%However, existing complex KBQA datasets and techniques still have two major limitations: 1) 
current KBQA datasets such as ComplexQuestions~\cite{coling_constraint_based} and ComplexWebQuestions~\cite{ComplexWebQuestions} mostly focus on multi-hop questions and questions with various constraints, while many question types in real-world KBQA applications are still not covered:
e.g., most existing datasets and models do not cover boolean questions (e.g. ``\textit{Was Stevie Nicks a musical composer?}'') and questions with multiple intentions (e.g. ``\textit{What is the job title of Gregory VII and when did he start working?}'').
Second, most existing KBQA models are not generalizable enough to process different types of questions. For example, information retrieval-based methods~\cite{Information_Extraction,MCCNN_Acl} attempt to rank candidate answers with respect to the given question, and therefore cannot answer boolean and dual-intention questions. The staged query graph generation methods~\cite{subgraph_embedding,STAGG,ACL_multi-hop} aim at generating the query graph and the answer nodes representing final answers. However, as the generation process is base on true relations in KB, these methods cannot effectively answer boolean questions since the corresponding query graph may be inaccurate.

In this paper, we construct the CLC-QuAD dataset to address these challanges. CLC-QuAD is the first Chinese large scale KBQA dataset over Wikidata that covers a wide variety of questions types, obtained by translating, verifying, and filtering LC-QuAD 2.0. We also propose a text-to-SPARQL model to 
handle the additional complexities in CLC-QuAD. In particular, we translate questions into SPARQL queries directly, motivated by prior research in semantic parsing~\cite{edit-sql, RAT-SQL, emnlp20lin}. In so doing, we avoid the difficulty of processing multi-type questions. We analyze the results and showcase the challenges brought by the added question types and also identify the unique challenges faced by Chinese KBQA.

We summarize our main contributions as follows:
\begin{itemize}
    \item We construct the first Chinese complex KBQA dataset (CLC-QuAD) based on Wikidata, which consists of multi-hop questions, dual-intent question, boolean questions and counting questions, shown in Figure~\ref{tab:dataset_sample}. We hope the release of the collected dataset would push forward Chinese KBQA research.
    \item We propose an end-to-end text-to-SPARQL baseline, which can effectively answer multi-type complex questions, such as fact questions, dual-intent questions, boolean questions and counting questions, with Wikidata as the background knowledge base. 
\end{itemize}

\begin{table}
\centering
\begin{tabular}[width=0.5\textwidth]{l}
\hline
{\fontsize{9pt}{0em} \selectfont 
\textbf{Sample 1: boolean question}
\par}\\
\hline
{\fontsize{9pt}{0em} \selectfont 
\textbf{English Question} 
\par} \\
{\fontsize{9pt}{0em} \selectfont 
Was Stevie Nicks a musical composer? \par} \\
% {\fontsize{10pt}{5em} \selectfont
% for Cloud computing? \par} \\
{\fontsize{9pt}{0em} \selectfont 
\textbf{Corresponding Chinese Question}
\par} \\
{\fontsize{9pt}{0em} \selectfont 史蒂薇·尼克斯是音乐作曲家吗?
\par} \\ 
{\fontsize{9pt}{0em} \selectfont
\textbf{Wikidata SPARQL} 
\par} \\
{\fontsize{8pt}{0em} \selectfont
ASK WHERE \{ wd:Q234691 wdt:P101 wd:Q207628 \}
\par} \\
\hline
{\fontsize{9pt}{0em} \selectfont
\textbf{Sample 2: dual intentions question}
\par} \\
\hline
{\fontsize{9pt}{0em} \selectfont
\textbf{English Question}
\par} \\
{\fontsize{9pt}{0em} \selectfont
What is the job title of Gregory VII and when\par} \\
{\fontsize{9pt}{0em} \selectfont
did he start working? \par} \\
{\fontsize{9pt}{0em} \selectfont
\textbf{Corresponding Chinese Question}
\par} \\
{\fontsize{9pt}{0em} \selectfont
格列高利七世的担任职位是什么,是从什么\par } \\
{\fontsize{9pt}{0em} \selectfont
时候开始的? \par} \\
{\fontsize{9pt}{0em} \selectfont
\textbf{Wikidata SPARQL}
\par} \\
{\fontsize{8pt}{0em} \selectfont
SELECT ?value1 ?obj WHERE \{ wd:Q133063 p:P39 ?s . \par} \\
{\fontsize{8pt}{0em} \selectfont
?s ps:P39 ?obj . ?s pq:P580 ?value1 \} \par} \\
\hline
\end{tabular}
\caption{Our corpus contains different types of questions in Chinese, including multi-hop questions, dual-intention questions, fact questions and boolean questions.}
\label{tab:dataset_sample}
\end{table}

\section{Related Work}

\subsection{KBQA Datasets}
Earlier KBQA datasets, such as Free917~\cite{FREE917} that contains 917 pairs of question and formal query in FREEBASE, are too small to train neural networks models. The
larger scale dataset WebQuestions~\cite{WebQuestions} consists of natural language question-answer pairs only, without formal queries. WebQuestionsSP~\cite{WebQuestionsSP} extracts part of the WebQuestions dataset and supplements it with the corresponding formal query and improves the multi-relation questions. SimpleQuestions~\cite{Simple_Question} builds a large-scale dataset, but only has one type of relation. Recently, ComplexWebQuestions~\cite{ComplexWebQuestions} presents a complex dataset that covers more components in the SPARQL grammar, but it only covers limited types of questions. LC-QUAD \cite{LC-QUAD} is the first complex KBQA dataset based on DBpedia. It starts by generating formal queries for DBpedia and then converts these template-based questions into natural language questions. 
QALD-9~\cite{QALD-9} is another well-known KBQA dataset based on DBpedia, which has more complex and colloquial questions than LC-QUAD. 
LC-QuAD 2.0~\cite{LC-QUAD2.0} is created in 2019, which expanded the number of data and increased formal query types.

\subsection{KBQA Systems}
In earlier work, ~\newcite{subgraph_embedding} starts to use neural networks to train question and answer vector representation, calculating the matching scores between question and candidate answers. ~\newcite{MCCNN_Acl, ACL_CATT_KBQA} uses CNN and attention to learn the relation and type features and obtains some improvement. These methods are based on information retrieval provide a simple solution by strengthening the connection between the question and the answer. But their candidate answer space can be huge for complex question. By contrast, semantic parsing methods can achieve much better score in complex KBQA dataset. ~\newcite{STAGG} puts forward a query graph structure, mapping the semantic parsing process to generate query graph process, which defines several operations to extend, including entity linking, attaching constraints, attribute recognition and so on. Following this idea, ~\newcite{coling_constraint_based, improve_relation_detect} adds new type constraints as well as dominant and recessive time constraints to solve more complex questions. Recently, \newcite{emnlp_state-transition, enhance_kv_memory, AQGnet, ACL_multi-hop} continues this line of research by defining new abstract query graph and designing stronger query graph representation to further
improve predicted query graph accuracy.

\section{Corpus Construction}

We translate all effective English questions in the LC-QuAD 2.0 dataset into Chinese. This work is distributed to 20 computer science students, supervised by 3 NLP researchers. Each item in LC-QuAD 2.0 has three paraphrased versions of questions, which provides more natural language variations for models to learn from and avoid over-fitting. We split three questions of each item and mix all questions to translate in order to guarantee Chinese question variations. Each question is first translated by one student, and then cross-checked and corrected by another student. Finally, a third student is in charge of verifying the original and corrected versions. As for the Chinese knowledge graph, we rely on Chinese descriptions in Wikidata. 

In addition, we also double check the gold SPARQL queries in LC-QuAD 2.0 and correct mistakes whenever we can. We check the correctness of the SPARQL queries from two aspects. First, we check all SPARQL queries in Wikidata Query Service to find syntax errors and get the answer labels at the same time. Second, we read both questions and SPARQL queries to make sure that questions are matched with queries. 
% Our dataset CLC-QuAD will be released on Github after the paper published. 

\begin{figure}[t]
    \centering
    \includegraphics[width=0.35\textwidth]{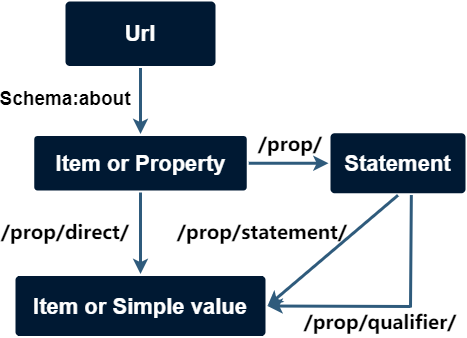}
    \caption{Wikidata prefixes: In our dataset, these prefix URLs are used to describe a single item which are a part of Wikidata prefix URLs. In Wikidata, the prefixes are used in RDF formats that allow short prefixes (such as Turtle and RDF). In order to execute the SPARQL query on Wikidata correctly, algorithms are required to generate both the right item and its right prefix.}
    \label{fig:wikibase}
\end{figure} 

\begin{table}[t]
\centering
\begin{tabular}[width=0.5\textwidth]{l l l l}
\hline
{\fontsize{9pt}{0em} \selectfont
\textbf{ } \par} & 
{\fontsize{9pt}{0em} \selectfont
\textbf{Freebase} \par} & 
{\fontsize{9pt}{0em} \selectfont
\textbf{DBpedia} \par} & 
{\fontsize{9pt}{0em} \selectfont
\textbf{Wikidata}  \par}
\\
\hline
{\fontsize{9pt}{0em} \selectfont \textbf{Entities}\par} &
{\fontsize{9pt}{0em} \selectfont 41 million\par} &
{\fontsize{9pt}{0em} \selectfont 6.6 million\par} &
{\fontsize{9pt}{0em} \selectfont 93 million\par} \\
{\fontsize{9pt}{0em} \selectfont \textbf{Triples}\par} &
{\fontsize{9pt}{0em} \selectfont 596 million\par} &
{\fontsize{9pt}{0em} \selectfont 13 billion\par} &
{\fontsize{9pt}{0em} \selectfont 13.9 billion\par} \\
{\fontsize{9pt}{0em} \selectfont \textbf{Relations}\par} &
{\fontsize{9pt}{0em} \selectfont 19456\par} &
{\fontsize{9pt}{0em} \selectfont 10000\par} &
{\fontsize{9pt}{0em} \selectfont 40276\par} \\
{\fontsize{9pt}{0em} \selectfont \textbf{Size (GB)}\par} &
{\fontsize{9pt}{0em} \selectfont 56.9\par} &
{\fontsize{9pt}{0em} \selectfont 9.25\par} &
{\fontsize{9pt}{0em} \selectfont 2030\par} \\
\hline
\end{tabular}
\caption{Statistics of main knowledge graphs from number of entities, number of triples, number of relations and size of knowledge graphs.}
\label{tab:Comparation_kg}
\end{table}

\begin{table*}[ptb]
\centering
\begin{tabular}{c c c c c c}
\hline
\textbf{Data Set} & 
\textbf{Size} & 
\textbf{Variation} & 
\textbf{Target KG} &
\textbf{formal query} &
\textbf{language}
\\
\hline
{FREE917} & {917} & {low} & {Freebase} & {yes} & {English} \\
{WebQuestions} & {5810} & {low} & {Freebase} & {no} & {English}\\
{WebQuestionsSP} & {4737} & {medium} & {Freebase} & {yes} & {English} \\
{SimpleQuestions} & {100k} & {low} & {Freebase} & {no} & {English} \\
{ComplexWebQuestions} & {34K} & {medium} & {Freebase} & {yes} & {English} \\
{LC-QuAD} & {5K} & {medium} & {DBpedia} & {yes} & {English}\\
{QALD-9} & {350} & {high} & {DBpedia} & {yes} & {English} \\
{LC-QuAD 2.0} & {30K} & {high} & {Wikidata, DBpedia} & {yes} & {English} \\
\hline
{CLC-QuAD(Ours)} & {28k} & {high} & {Wikidata} & {yes} & {Chinese} \\
\hline
\end{tabular}
\caption{A comparison of existing datasets having questions and corresponding formal queries}
\label{tab:compare_with_other_dataset}
\end{table*}

\subsection{Knowledge Graph Statistics}
% TODO
We show the statistics of three knowledge graphs in Table~\ref{tab:Comparation_kg}. Freebase~\cite{FREEBASE} is a collaboratively edited knowledge base designed to be a public repository of the world’s knowledge. DBpedia~\cite{DBpedia} is a knowledge graph mainly based on the English Wikipedia. Wikidata~\cite{Wikidata} is a free, open, and massively linked knowledge base. Compared with above two knowledge graphs, we can find that Wikidata contains a larger number of entities and relations. More importantly, Wikidata defines prefixes to describe the IRIs of the RDF resources, that are suitable for variety of SQARQL queries, shown in Figure~\ref{fig:wikibase}. \\

\subsection{Data Statistics and Analysis}

We compute the statistics of both LC-QuAD 2.0 and CLC-QuAD, and carry out a data analysis focusing on semantic coverage and question types. In this section, we will also compare them with other complex knowledge base question answering datasets.\\
\\
\textbf{Data statistics} Table~\ref{tab:compare_with_other_dataset} summarizes the statistics of nine datasets. CLC-QuAD contains 28k+ pairs of question and SPARQL query in total, which is comparable to or bigger than most commonly used KBQA datasets. As for dataset variation, FREE917, WebQuestions and SimpleQuestion focus on simple questions without constraints, so most of data in these datasets even don't have corresponding formal queries. LC-QuAD contains more question types and its SPARQL components cover \textsf{SELECT}, \textsf{COUNT}, \textsf{ASK} and \textsf{DISTINCT}. ComplexWebQuestions bulids a complex dataset by adding different and complicated constraints and its SPARQL components cover \textsf{SELECT}, \textsf{DISTINCT}, \textsf{FILTER}, \textsf{LANG}, \textsf{DATATIME} etc. In LC-QuAD 2.0 and CLC-QuAD, we ensure that corpus contains enough examples for all common SPARQL patterns in order to describe the question correctly. CLC-QuAD covers all the following SPARQL components: \textsf{SELECT} with one or multiple variable, \textsf{COUNT}, \textsf{ASK}, \textsf{DISTINCT}, \textsf{FILTER}, \textsf{CONTAINS}, \textsf{YEAR}, \textsf{STRSTARTS}, \textsf{LIMIT}, \textsf{ORDER BY}, \textsf{LANG}. Noticeably, most of datasets are built on Freebase and DBpedia, which makes datasets based on Wikidata more distinctive and challenging, as researchers need to explore the connection between questions and the structure of the Wikidata knowledge graph. In addition, CLC-QuAD provides Chinese questions that help study KBQA in a cross-lingual setting. Since the same question can be expressed quite differently in Chinese and in English, we also present statistics of characters for each language.
\begin{table}[t]
\centering
\begin{tabular}[width=0.5\textwidth]{c c c c c}
\hline
{\fontsize{9pt}{0em} \selectfont
{ }\par} & 
{\fontsize{9pt}{0em} \selectfont
\textbf{CLC}\par} & 
{\fontsize{9pt}{0em} \selectfont
{LC2.0}\par} & 
{\fontsize{9pt}{0em} \selectfont
{CWQ}\par} &
{\fontsize{9pt}{0em} \selectfont
{LC}\par}
\\
\hline
{\fontsize{9pt}{0em} \selectfont{\# Question}\par} & 
{\fontsize{9pt}{0em} \selectfont{28,409}\par} &
{\fontsize{9pt}{0em} \selectfont{30,226}\par} &
{\fontsize{9pt}{0em} \selectfont{34,689}\par} &
{\fontsize{9pt}{0em} \selectfont{5,000}\par}\\
{\fontsize{9pt}{0em} \selectfont{Avg.\# Q len}\par} & 
{\fontsize{9pt}{0em} \selectfont{20.1}\par} &
{\fontsize{9pt}{0em} \selectfont{10.7}\par} &
{\fontsize{9pt}{0em} \selectfont{13.4}\par} &
{\fontsize{9pt}{0em} \selectfont{11.4}\par}\\
{\fontsize{9pt}{0em} \selectfont{\# Vocab}\par} & 
{\fontsize{9pt}{0em} \selectfont{32,683}\par} &
{\fontsize{9pt}{0em} \selectfont{45,476}\par} &
{\fontsize{9pt}{0em} \selectfont{30,627}\par} &
{\fontsize{9pt}{0em} \selectfont{9,682}\par}\\
{\fontsize{9pt}{0em} \selectfont{\# Entities}\par} & 
{\fontsize{9pt}{0em} \selectfont{20,577}\par} &
{\fontsize{9pt}{0em} \selectfont{21,485}\par} &
{\fontsize{9pt}{0em} \selectfont{12,500}\par} &
{\fontsize{9pt}{0em} \selectfont{3,968}\par}\\
{\fontsize{9pt}{0em} \selectfont{\# Relations}\par} & 
{\fontsize{9pt}{0em} \selectfont{3,447}\par} &
{\fontsize{9pt}{0em} \selectfont{3,660}\par} &
{\fontsize{9pt}{0em} \selectfont{825}\par} &
{\fontsize{9pt}{0em} \selectfont{748}\par}\\
{\fontsize{9pt}{0em} \selectfont{\# Keyword}\par} & 
{\fontsize{9pt}{0em} \selectfont{11}\par} &
{\fontsize{9pt}{0em} \selectfont{11}\par} &
{\fontsize{9pt}{0em} \selectfont{6}\par} &
{\fontsize{9pt}{0em} \selectfont{4}\par}\\
{\fontsize{9pt}{0em} \selectfont{Dual Intent}\par} & 
{\fontsize{9pt}{0em} \selectfont{\Checkmark}\par} &
{\fontsize{9pt}{0em} \selectfont{\Checkmark}\par} &
{\fontsize{9pt}{0em} \selectfont{\XSolidBrush}\par} &
{\fontsize{9pt}{0em} \selectfont{\XSolidBrush}\par}\\
{\fontsize{9pt}{0em} \selectfont{Boolean Intent}\par} & 
{\fontsize{9pt}{0em} \selectfont{\Checkmark}\par} &
{\fontsize{9pt}{0em} \selectfont{\Checkmark}\par} &
{\fontsize{9pt}{0em} \selectfont{\XSolidBrush}\par} &
{\fontsize{9pt}{0em} \selectfont{\Checkmark}\par}\\
{\fontsize{9pt}{0em} \selectfont{Constraint}\par} & 
{\fontsize{9pt}{0em} \selectfont{\Checkmark}\par} &
{\fontsize{9pt}{0em} \selectfont{\Checkmark}\par} &
{\fontsize{9pt}{0em} \selectfont{\Checkmark}\par} &
{\fontsize{9pt}{0em} \selectfont{\XSolidBrush}\par}\\
\hline
\end{tabular}
\caption{Statistics of knowledge base question answering with more details. The number are counted over the entire datasets. For CLC-QuAD, we use professional Chinese word segmentation tool to compute the number of vocab.}
\label{tab:Semantic_Coverage}
\end{table}
\begin{figure}[t]
    \centering
    \includegraphics[width=0.5\textwidth]{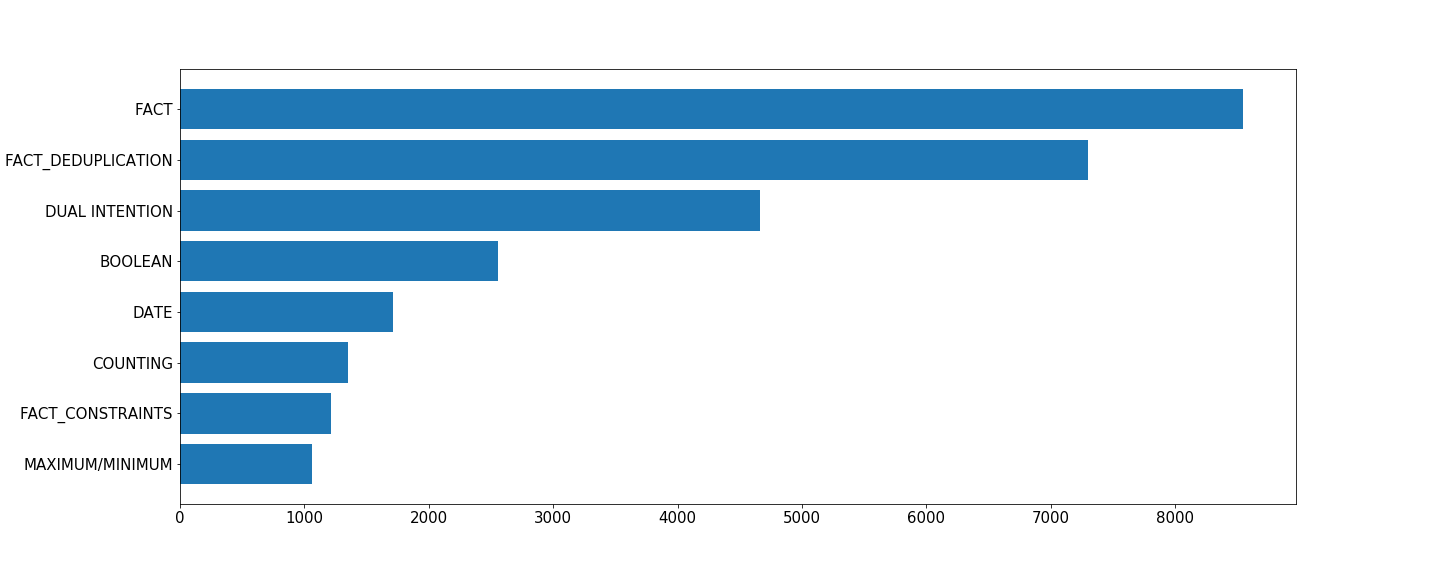}
    \caption{Distributions of question types in CLC-QuAD.}
    \label{fig:question_distribution}
\end{figure}
\\
\\
\textbf{Semantic Coverage} As shown in Table~\ref{tab:Semantic_Coverage}, we make a statistical comparison against previous KBQA datasets in this task. Obviously, CLC-QuAD and LC-QuAD 2.0 is vast in coverage of knowledge graph entities and relations, which can enable better generalization in model training. In addition, the SPARQL queries in CLC-QuAD and LC-QuAD 2.0 cover all common SPARQL keywords. As for vocabulary in datasets, we believe our translating work is much better than machine translation, as the translators make their focus on both questions and correct SPARQL queries so that they can use suitable word in different situations, as demonstrated by the vocabulary used in our dataset.
\\
\\
\textbf{Question Distribution} As shown in Figure \ref{fig:question_distribution}, CLC-QuAD contains a fairly diverse set of question types over knowlegde graphs. Unsurprisingly, FACT and FACT\_DEDUPLICATION questions are the two most commonly seen in KBQA systems. Among the rest of question types, approximately 35\% are DUAL INTENTION questions, which pose huge challenges to semantic parsing. Another 20\% of this subgroup is BOOLEAN question, which is also very common in real-world applications, but is often ignored by researchers. BOOLEAN questions can be hard to handle traditional algorithms, because it is more like a classification problem rather than mapping to existing knowledge graphs. Also, DATE, COUNTING, MAX/MIN questions strengthen the diversity of entire dataset and involve more corresponding SPARQL components.

\section{Approach}

In this section, we describe our text-to-SPARQL model 
in two parts: relation-aware attention and multi-types pointer network.
Figure~\ref{fig:model_fig} illustrates the overview of the model.

\subsection{Relation-aware Attention Encoder}

As for the embedding layer, we consider two options as input to the next layer. 
First choice is pretrained word embedding~\cite{tencent}. 
Alternatively, we use BERT~\cite{bert} as the initial representations of the word.
Formally, we concatenate question $Q$ and all entities $E$ and relations $R$ by this structure.
$$[CLS],q_1...,[SEP],e_1,[SEP]...,r_{|R|},[SEP] $$
This sequence is fed into the pretrained BERT model and use the last hidden states as the initial representations.

To capture graph information across the question, we use relation-aware self attention layers~\cite{relation_aware_attention} to compute new contextual representations of questions, entities and relations item. 
We define the input $X = \{x_i\}_{i=1}^n$ where $x_i \in \{q_i...,e_i...,r_i\} = Q \cup E \cup R$.
This is the relation-aware self-attention process for each layer (consisting of $H$ heads):
\begin{equation}
 \begin{split}
e_{ij}^{(h)}&= \frac{x_i W_Q^{(h)} (x_j W_K^{(h)} + r_{ij}^K)^\mathrm{T} }{\sqrt{d_z/H}}\\
\alpha^{(h)}&=Softmax(e^{(h)})\\
c_i^{(h)}&=\sum_{j=1}^n\alpha_{ij}^{(h)}(x_j W_V^{(h)} + r_{ij}^V)
\end{split}
\end{equation}
where $W_Q^{(h)}, W_K^{(h)}, W_V^{(h)} \in \mathbb{R}^{d_x\times(d_x/H)}$, $r^K = r^V \in \mathbb{R}^{n\times n\times (d_x/H)}$. 
$r_{ij}$ is the vector which represent the relation type between the two item $x_i$ and $x_j$ in the input. 
And in our model, $r_{ij}$ is designed as a learned parameter for edge types in graph, such as <wdt-wd> and <wdt-p>. 
After the attention procedure, we use fully connected feed-forward networks to transform the attention output and 
the ReLU activation is used between the two fully connected networks.

In sum, every relation-aware self attention layer use the corresponding graph $G_Q$ and compute a new contextual representations of question word, entities and relations. 

\subsection{Pointer Network Decoder}

Because the output SPARQL query consists of entities, relations and SPARQL keywords, 
we use pointer networks with three separate independent scaled dot-product attention for different components.
During the decoding process, we use Long Short Term Memory (LSTM) with attention to generate SPARQL queries by incorporating the representation of entities, relations and SPARQL keywords.

Denote the decoding step as $t$, we provide the decoder input as a concatenation of the embedding of the SPARQL query token $S_{t}$ and the context vector $c_{t}$:
\begin{equation}
h_{t+1} = \mathrm{LSTM}([S_t; c_t], h_t)
\label{decode_lstm}
\end{equation}
where $h_{t}$ is the hidden state of the decoder $\mathrm{LSTM}$ and the hidden state $h_0$ is initialized as random, as well as $S_t, c_t$. And $S_t$ is generated by the types of the SPARQL query token. 
$$
S_t=\begin{cases}
	W_{k}h^{keyword} + b_{k} & \text{token $\in$ Keyword}\\
	W_{e}h^{entity} + b_{e} & \text{token $\in$ Entity}\\
	W_{r}h^{relation} + b_{r} & \text{token $\in$ Relation}\\
	\end{cases}
$$
where $h^{entity}, h^{relation}$ are extracted from encoder result $h^{enc}$ and $h^{keyword}$ is a learnable embedding. In addition, we compute $c_t$ by multi-head attention with $combined$ components representation as follows:
\begin{equation}
 \begin{split}
 h_i &\in [h^{enc};h^{keyword}]\\
 e_i^{cb}&= \frac{h_t W_{Q}^{cb} (h_i W_K^{cb})^\mathrm{T} }{\sqrt{d_h}}\\
\alpha^{cb}&=Softmax(e^{cb})\\
c_t &=\sum_{i=1}\alpha_{i}^{cb} (h_i W_V^{cb})
 \end{split}
\end{equation}
where $h_t$ is equal to $h_t$ in Equation~\ref{decode_lstm} and $d_h$ is the dimension of $h_t$. The context vector $c_t$ consists of attentions to both the question, entities, relations and SPARQL keywords for current step $k$.

As for output layer, our decoder is designed to generate an entity, a relation or a SPARQL keyword (eg. \textsf{COUNT}, \textsf{ASK}, \textsf{FILTER}, \textsf{ORDER BY}). In addition, entities and relations will be changeable based on different candidate graph so that we choose three separate independent layer for different components and then use softmax function to compute the output probability distribution:
\begin{equation}
 \begin{split}
 o_t^{k} &= \frac{h^{keyword}([h_t;c_t]W_{k}^{o})}{\sqrt{d_h}}\\ 
 o_t^{e} &= \frac{h^{entity}([h_t;c_t]W_{e}^{o})}{\sqrt{d_h}}\\ 
 o_t^{r} &= \frac{h^{relation}([h_t;c_t]W_{r}^{o})}{\sqrt{d_h}}\\ 
 P_t &= softmax([o_t^{k};o_t^{e};o_t^{r}])\\
 \end{split}
\end{equation}

As for loss computing, we compute the output character with ground truth SPARQL query character by the cross entropy loss function.

\begin{table*}[t]
\centering
\begin{tabular}[width=0.5\textwidth]{l c c| c c}
\hline
{\fontsize{9pt}{0em} \selectfont
\textbf{}\par} & 
\multicolumn{2}{c|}{\fontsize{9pt}{0em} \selectfont
\textbf{LC-QuAD 2.0}\par} &
\multicolumn{2}{c}{\fontsize{9pt}{0em} \selectfont
\textbf{CLC-QuAD}\par}\\
{\fontsize{9pt}{0em} \selectfont
\textbf{}\par} & 
{\fontsize{9pt}{0em} \selectfont
\textbf{answer F1}\par} &
{\fontsize{9pt}{0em} \selectfont
\textbf{query match}\par} &
{\fontsize{9pt}{0em} \selectfont
\textbf{answer F1}\par} &
{\fontsize{9pt}{0em} \selectfont
\textbf{query match}\par} \\
\hline
{\fontsize{9pt}{0em} \selectfont AQG-net~\cite{AQGnet}\par} & 
{\fontsize{9pt}{0em} \selectfont 44.9\par} &
{\fontsize{9pt}{0em} \selectfont 37.4\par} &
{\fontsize{9pt}{0em} \selectfont 38.5\par} &
{\fontsize{9pt}{0em} \selectfont 32.1\par}\\
{\fontsize{9pt}{0em} \selectfont Multi-hop QGG ~\cite{ACL_multi-hop}\par} & 
{\fontsize{9pt}{0em} \selectfont 52.6\par} &
{\fontsize{9pt}{0em} \selectfont 43.2\par} &
{\fontsize{9pt}{0em} \selectfont 46.5\par} &
{\fontsize{9pt}{0em} \selectfont 39.7\par}\\
\hline
{\fontsize{9pt}{0em} \selectfont Our approach + Tencent Word\par} &
{\fontsize{9pt}{0em} \selectfont 52.9\par} &
{\fontsize{9pt}{0em} \selectfont 48.4\par} &
{\fontsize{9pt}{0em} \selectfont 45.6\par} &
{\fontsize{9pt}{0em} \selectfont 40.2\par}\\
{\fontsize{9pt}{0em} \selectfont Our approach + Bert\par} & 
{\fontsize{9pt}{0em} \selectfont \textbf{59.3}\par} &
{\fontsize{9pt}{0em} \selectfont \textbf{55.4}\par} &
{\fontsize{9pt}{0em} \selectfont \textbf{51.8}\par} &
{\fontsize{9pt}{0em} \selectfont \textbf{45.4}\par}\\
{\fontsize{9pt}{0em} \selectfont { }w/o graph relation-aware self-attention \par} & 
{\fontsize{9pt}{0em} \selectfont 50.1\par} &
{\fontsize{9pt}{0em} \selectfont 46.6\par} &
{\fontsize{9pt}{0em} \selectfont 42.0\par} &
{\fontsize{9pt}{0em} \selectfont 36.7\par}\\
{\fontsize{9pt}{0em} \selectfont { }w/o decoder separate attention \par} & 
{\fontsize{9pt}{0em} \selectfont 55.5\par} &
{\fontsize{9pt}{0em} \selectfont 51.2\par} &
{\fontsize{9pt}{0em} \selectfont 48.9\par} &
{\fontsize{9pt}{0em} \selectfont 42.7\par}\\
\hline
\end{tabular}
\caption{Performance of various methods over all answers and all queries on both LC-QuAD 2.0 and CLC-QuAD. }
\label{tab:baseline_on_two_datasets}
\end{table*}

\begin{table*}[t]
\centering
\begin{tabular}{l c c c c c c}
\hline
{\fontsize{9pt}{0em} \selectfont
\textbf{Split}\par} & 
{\fontsize{9pt}{0em} \selectfont
\textbf{Dual}\par} &
{\fontsize{9pt}{0em} \selectfont
\textbf{Boolean}\par} &
{\fontsize{9pt}{0em} \selectfont
\textbf{Fact}\par} & 
{\fontsize{9pt}{0em} \selectfont
\textbf{Max/Minimum}\par} & 
{\fontsize{9pt}{0em} \selectfont
\textbf{COUNTING}\par} &
{\fontsize{9pt}{0em} \selectfont
\textbf{Qualifier}\par} \\
\hline
{\fontsize{9pt}{0em} \selectfont AQG-net~\cite{AQGnet}\par} & 
{\fontsize{9pt}{0em} \selectfont -\par} &
{\fontsize{9pt}{0em} \selectfont 50.6\par} &
{\fontsize{9pt}{0em} \selectfont 34.3\par} &
{\fontsize{9pt}{0em} \selectfont 41.2\par} &
{\fontsize{9pt}{0em} \selectfont 25.8\par} &
{\fontsize{9pt}{0em} \selectfont 17.2\par} \\
{\fontsize{9pt}{0em} \selectfont Multi-hop QGG~\cite{ACL_multi-hop}\par} &
{\fontsize{9pt}{0em} \selectfont -\par} &
{\fontsize{9pt}{0em} \selectfont -\par} &
{\fontsize{9pt}{0em} \selectfont 42.3\par} &
{\fontsize{9pt}{0em} \selectfont 45.4\par} &
{\fontsize{9pt}{0em} \selectfont -\par} &
{\fontsize{9pt}{0em} \selectfont \textbf{25.6}\par} \\
\hline
{\fontsize{9pt}{0em} \selectfont \textbf{Our approach}\par} & 
{\fontsize{9pt}{0em} \selectfont \textbf{51.3}\par} &
{\fontsize{9pt}{0em} \selectfont \textbf{55.3}\par} &
{\fontsize{9pt}{0em} \selectfont \textbf{46.6}\par} &
{\fontsize{9pt}{0em} \selectfont \textbf{47.1}\par} &
{\fontsize{9pt}{0em} \selectfont \textbf{30.1}\par} &
{\fontsize{9pt}{0em} \selectfont 18.8\par} \\
\hline
\end{tabular}
\caption{Performance of various question types on CLC-QuAD. Some item is empty which represent that model can not deal with such type questions. }
\label{tab:question_type_result}
\end{table*}

\section{Experimental Results}
We implemented our model in PyTorch. We use pretrained model BERT~\cite{bert} as the embedding layer and set $h^{keyword}=256, h^{entity}=h^{relation}=768$. We use 6 relation-aware attention layers to capture graph information. Within attention layers, the hidden size $d_z$ is set as 256 and the number of head is 8. And we use 2 LSTM layers with 0.2 dropout for decoder and all hidden size $h_t$ is 512. As for dropout rate, we set 0.1 for all attention layers and 0.2 for LSTM layers. To show that the effectiveness of our model is not mainly due to the use of the pre-trained model, we also experiment replacing BERT with the 200-dimensional Chinese word embedding~\cite{tencent}.

We used the Adam optimizer~\cite{adam} with the default hyperparameters. During the first 2 epochs, the model learning rate linearly increases from 0 to $1\times 10^{-3}$. Afterwards, it will be multiplied by 0.8 if the validation loss increases compared with the previous epoch. We use a batch size of 16 and train for up to 15 epochs. When using BERT, we use a separate constant learning rate of $3\times 10^{-6}$ to fine-tune it, a batch size of 4 and train for up to 25 epochs.

\subsection{Baselines}

Because there is no model aiming at multi-type KBQA datasets, we modify two state-of-the-art models based on earlier datasets for our model to compare to. Due to limitation of their models, some types of questions cannot be handled, so we test them with specific types of questions.\\
\textbf{AQG-net} proposed by \newcite{AQGnet} first uses a neural network-based generative model to generate an abstract query graphs that describes logical query structures, filling up it with all possible candidate permutations and then utilizes existing ranking model to get the most suitable query. In our experiment, we redesign the abstract query graphs due to the question types and relations are different and more complicated. \\
\textbf{Multi-hop QGG} proposed by \newcite{ACL_multi-hop} explores a new strategy to expand the candidate query graph with both constraints and core paths. And it applies the REINFORCE algorithm to learn a policy function by using the F1 score of the predicted answers with respect to the ground truth answers as reward. In our experiment, we totally redefine the way of extending the current SPARQL path and adapt the model features to meet the our dataset.

\subsection{Evaluation Metrics}

We evaluate models from two aspects. First, we use the F1 scores metrics to calculate the accuracy between ground truth answer and answers obtained from predicted SPARQL queries. Second, as semantic parsing methods model aim to predict correct SPARQL queries by question analysis, the answer accuracy can not represent its analytical capability. Instead of simply taking string match, we decompose predicted SPARQL queries into different triples such as $(?ans_1,wdt$:$P31,wd$:$Q22675015)$, $(?var_1, order\ by, ascend)$ and compute scores for the triples set using exact set match, which measures whether the predicted query is entirely equivalent to the gold query. The predicted SPARQL query is correct only if all of the components are right. Because we employ a triples set comparison, this exact matching metric is invariant to the order of the components.

\subsection{Overall Result}

\textbf{LC-QuAD 2.0} We report the overall results of our approach and others on LC-QuAD 2.0 in Table ~\ref{tab:baseline_on_two_datasets}. Our method achieves the performance of 55.4\% query exact match scores and 59.3\% answer F1 scores, which is better than other methods. This demonstrates the effectiveness of our approach and that the text-to-SPARQL method can handle the semantics of multi-type questions to generate complex SPARQL queries. Furthermore, from the ablation study, we find that model without BERT has 7.0\% decline but still achieves state-of-the-art. This shows that the effectiveness of our model is not simply because of the use of BERT. We also
test two components of our method. Without the relation-aware self attention, the score is nearly 9\% lower and it shows that the information of knowledge graph is very important to the model and the relation-aware self attention can effectively encode the items in the graph. Without the separate attention in pointer networks, the model has a little drop in results.\\
\textbf{CLC-QuAD} Table~\ref{tab:baseline_on_two_datasets} also shows the results of models in our new dataset CLC-QuAD. 
Similar to LC-QuAD 2.0, our model with BERT achieves 51.8\% answer F1 score and 45.4\% query exact matching accuracy.
In addition, we find that all models get pretty weak scores compared with results in LC-QuAD 2.0. In the results of our approach with BERT, there is 10\% decline which is a huge gap. This demonstrates that the Chinese representation is still a big challenge for model semantic parsing. 

In addition, we split our dataset into six question types. Table~\ref{tab:question_type_result} shows the performance of various question types on CLC-QuAD. We find that there is a huge gap in accuracy between different types of questions. We find that Counting and Qualifier questions are harder due to more complex semantics, which brings challenges to semantic parsing. Our approach achieves 51.3\%, 55.3\% and 47.1\% accuracy in Dual, Boolean and Max/Minimum questions relatively, but the accuracy for Qualifier and Counting questions is only 18.8\% and 30.1\%. As for other methods, Multi-hop QGG is designed for complex questions with constraints and it achieves best performance in Qualifier questions, but it cannot handle Dual, Boolean and Counting questions. AQG-net is designed to generate abstract query graphs, which can answer most of types questions, but the
performance is inferior to our proposed approach.  This also demonstrates our model is more competitive in answering different types of questions.

\section{Conclusion}

%TODO: rewrite conclusion
In this paper, we introduced CLC-QuAD, a large-scale dataset of Chinese Complex Knowledge base question answer dataset. The dataset features wide coverage with regard to question semantics and types. To improve semantic parsing research with large knowledge graph, our dataset provided the pair of a question and its corresponding SPARQL query based on Wikidata. In addition, we proposed a multi-type KBQA model by generating the SPARQL query directly. This approach achieves state of the art on LC-QuAD 2.0 and CLC-QuAD. 

% to fill up the blank baseline in Wikidata, we experimented with two competitive KBQA semantic parsing approaches on both LC-QuAD 2.0 and CLC-QuAD. The result is far from satisfactory and the both performance of two models shows that Chinese KBQA is more challenging. Our dataset can serve as a starting point for further research on this task, which can be helpful to the investigation of Chinese complex KBQA.

\section*{Acknowledgements}

This work was supported by Alibaba Group through Alibaba Innovative Research Program.

% This document has been adapted

% Entries for the entire Anthology, followed by custom entries
\bibliography{anthology,custom}
\bibliographystyle{acl_natbib}

\appendix

\end{CJK*}
\end{document}